\documentclass[]{fairmeta}

\usepackage{pifont}
\definecolor{checkgreen}{rgb}{0.13, 0.55, 0.13}
\definecolor{checkorange}{rgb}{0.93, 0.53, 0.18}
\newcommand{\yes}{\textcolor{checkgreen}{\ding{51}}}
\newcommand{\no}{\textcolor{red}{\ding{55}}}
\newcommand{\kinda}{\textcolor{checkorange}{$\boldsymbol{\sim}$}\xspace}

\usepackage[most]{tcolorbox}
\usepackage{enumitem}
\usepackage{booktabs}
\usepackage{array}
\usepackage{tabularx}

\title{Modality Maturity Index: A benchmark for assessing multimodal capabilities of omni models}

\author[1]{Rohit Patel}
\author[1]{Dieuwke Hupkes}
\author[1]{Sloan Strader}

\affiliation[1]{Meta Superintelligence Labs}

\date{May 20, 2026}
\correspondence{\email{mmipaper@rpeml.com} or \email{dieuwkehupkes@meta.com}}
\metadata[Code]{\url{https://github.com/facebookresearch/modality-maturity-index}}
\metadata[Dataset]{\url{https://huggingface.co/datasets/facebook/mmi}}

\abstract{
Frontier language models are increasingly marketed as multimodal, or even \emph{omni} systems that can perceive and respond across text, image, audio, video and document inputs and outputs.
Existing evaluation frameworks, however, focus almost exclusively on bimodal understanding, typically text plus one other modality.
Few benchmarks probe how models combine multiple non-text modalities within a single prompt and even fewer evaluate whether models choose the correct output modalities.
To address this gap, we propose the Modality Maturity Index (\bnshort), a benchmark designed to evaluate the multimodal capabilities of large language models across five modalities (text, image, audio, video and document) and combinations of up to three modalities in both inputs and outputs.
\bnshort consists of \nsamples questions, each carefully crafted to require the model to demonstrate its understanding of multiple input modalities and to generate responses that incorporate various output formats.
The questions are designed to be self-contained, with clear expectations for the correct modality or mix of modalities required for an accurate response.
Every \bnshort prompt carries human-authored rubric criteria for each output modality expected in the response; a model's \emph{\mmiscore} expresses the average of the per-modality scores for each prompt.
Because low scores can reflect either failure to generate a modality (lack of presence) or failure to generate correct content, we introduce also a supplementary \emph{\mpsfull} (\mps), a per-prompt F1 over the expected output modalities.
Applying \bnshort to five frontier multimodal models, we find that the \mps ranges from only 15.6 (Claude Opus 4.6) to 34.9 (GPT-5.4).
Given the low availability of returned modalities to even grade, we report \mps as our main result pending model improvements.
To assess the viability of judging output correctness with LLM judges and rubrics, we do run a separate experiment in which we give models access to tools generating images, audio and video.
On the assets that generates, we find that an LLM judge applying the rubrics agrees with \emph{rubric-blind} human annotators (who score the outputs directly and never see the criteria) on 70.8\% of judgments (Scott's $\pi$ of 0.41), ranging from 66.5\% on image to 74.4\% on video.
These findings highlight the need for further research into the development of more effective multimodal models as well as benchmarks to properly evaluate true multimodality.
}

\usepackage{xspace}
\newcommand{\benchmarkname}{Modality Maturity Index}
\newcommand{\bn}{\benchmarkname\xspace}
\newcommand{\bnshort}{MMI\xspace}
\newcommand{\mpsfull}{Modality Presence Score\xspace}
\newcommand{\mps}{MPS\xspace}
\newcommand{\mmiscore}{MMI Value\xspace}
\newcommand{\mmiscores}{MMI Values\xspace}
\newcommand{\nsamples}{893\xspace}

\begin{document}

\maketitle

\section{Introduction}
\label{sec:introduction}

\begin{figure}
    \centering
    \begin{subfigure}[c]{0.45\textwidth}
        \centering
        \includegraphics[width=\textwidth]{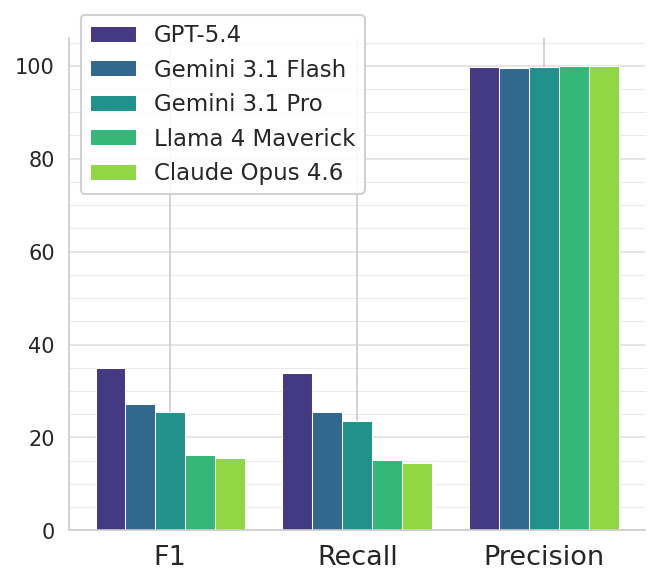}
    \caption{Per-model precision, recall and F1 (\mps) on \bnshort.}
    \label{fig:overall_rpf}
    \end{subfigure}
    \hfill
    \hfill
    \begin{subfigure}[c]{0.52\textwidth}
        \centering
        \begin{tabular}{lcccc}
\toprule

    & \rotatebox{90}{\textbf{Gemini 3}} & \rotatebox{90}{\textbf{Llama 4}} & \rotatebox{90}{\textbf{GPT-5.4}} & \rotatebox{90}{\textbf{Claude 4}} \\
\midrule
\scriptsize Text              & \yes   & \yes   & \yes   & \yes   \\
\scriptsize Image + Text      & \yes   & \yes   & \yes   & \yes   \\
\scriptsize Video + Text      & \yes   & \no    & --     & --     \\
\scriptsize Audio / Speech    & \yes   & --     & --     & --     \\
\scriptsize Cross-modal       & \no    & \no    & \no    & \no    \\
\bottomrule
\end{tabular}

        \centering
        \caption{Modality coverage of benchmarks reported in recent frontier model release notes.}
    \label{fig:benchmark_coverage}
    \end{subfigure}
    \caption{\textbf{Summary model performance and benchmark coverage of frontier models.} a) Average per-prompt precision, recall and F1 (the \mps) under lenient detection. F1 is low across the board, with even the best-performing model (GPT-5.4) not surpassing an F1 of 35 on returning the correct modality. b) Modality coverage of the evaluation benchmarks reported in recent frontier model release notes; no release reports cross-modal output benchmarks.}
    \label{fig:main_results}
\end{figure}

Recently released frontier LLMs are routinely marketed as being able to reason, plan, perceive and respond across multiple modalities \citep{gpt54,claude46,gemini31pro,musespark2026}.
However, the release reports of these same models scarcely cover all modalities in their benchmark tables: almost all of them omit one or more modalities, and none report results for benchmarks that explicitly require cross-modal output (see \cref{fig:benchmark_coverage}).
Furthermore, while there is a wealth of bimodal benchmarks assessing image or video understanding and to some extent generation \citep[see, e.g.][for overviews]{zhang-etal-2025-redundancy-principles,zhang-etal-2025-lmms}, mixed-modality and audio benchmarks \citep[e.g.][]{xie-etal-2025-mme-unify,li-etal-2025-omnibench} are rare, and benchmarks that require fully cross-modal processing, combining multiple non-text modalities in both input \emph{and} output, are virtually non-existent.

In this paper, we propose the \bn (\bnshort), the first benchmark we are aware of that directly assesses the multimodal capabilities of LLMs across multiple modalities and combinations of modalities, in both inputs \emph{and} outputs, and in particular the first to evaluate whether models \emph{choose the correct output modality} to respond to a self-contained user request.
\bnshort consists of \nsamples questions, carefully crafted to require understanding of various input modalities and to elicit responses that incorporate multiple output formats.
It encompasses five different modalities (text, image, audio, video and document) and spans prompts with up to three input and up to three output modalities.
The \bnshort prompts are designed to be self-contained, with clear expectations for the correct modality or mix of modalities.
All \bnshort prompts furthermore come with human-authored rubric criteria, each tied to one of the output modalities expected in the response.

\paragraph{Scoring MMI}
A prompt's \emph{\mmiscore} is the macro-average of the rubric scores over the output modalities expected in the response: we first average the criterion scores within each expected output modality, then average those per-modality means, so that every expected output modality counts equally regardless of how many criteria were written for it. 
The \mmiscore conflates two different failure modes: failing to return an asset for a particular modality, and returning an asset with the correct modality that does not answer the question correctly.
To separate these two cases, we also introduce a second score, the \emph{\mpsfull} (\mps), a per-prompt F1 over the expected output modalities, which measures production alone. 
The full definitions of both \mmiscore and \mps can be found in \cref{subsec:scoring}.

Testing a set of five multimodal models, namely GPT-5.4 \citep{gpt54}, Claude Opus 4.6 \citep{claude46}, Gemini 3.1 Pro \citep{gemini31pro}, Gemini 3.1 Flash \citep{gemini31flash}, and Llama 4 Maverick \citep{llama4}, we find that all of them struggle to both ingest modalities and return the right modalities at all.
The \mps ranges from only 15.6 (Claude Opus 4.6) to 34.9 (GPT-5.4): even ignoring answer correctness entirely, model performance is staggeringly poor.
As computing the \mmiscore on the few returned assets would not yield a meaningful extra signal, we report \mps as our main result, pending stronger future models.

To validate the rubrics, we instead run a second experiment, where we give three models access to tools they can use to generate images, videos and audio (\cref{subsec:rubric_eval_setup,subsec:rubric_evaluation}).
This raises modality presence substantially, yielding gradeable answers for 1{,}499 output modalities (71\%, or 76\% excluding documents, for which we did not provide a model tool).
We find that an LLM judge using the rubrics agrees with a human annotator on 70.8\% of the 1{,}499 judgments this yields (Scott's $\pi$ of 0.41), ranging from 66.5\% on image to 74.4\% on video.
Our analysis attributes the gap to judge errors, the difficulty of writing criteria for open-ended requests, and the difference between a rubric-driven judgment of a single modality and an unaided one made with the full response available (\cref{subsec:rubric_evaluation}).

\paragraph{Contributions}
In sum, our main contributions are:
\begin{itemize}[leftmargin=*,topsep=2pt,itemsep=1pt]
    \item \textbf{A cross-modal dataset.} \nsamples self-contained prompts spanning five modalities (text, image, audio, video and document), with up to three modalities in both input and output, and human-authored per-modality rubrics for every prompt. To our knowledge, this is the first dataset built to test whether a model chooses the correct output modality, and then assess the quality of generated assets.
    \item \textbf{An index over that dataset, and a diagnostic for it.} The \mmiscore grades a response against the prompt's rubrics. The \mps decomposes it, scoring modality presence as a per-prompt mean F1, complemented by precision, recall, pass rate and input failure rate. Both are implemented in the released harness.
    \item \textbf{An evaluation of five frontier models, and of the index itself.} The \mps ranges from 15.6 to 34.9, with failures predominantly recall-driven: audio is by far the hardest output modality, and most models default to text even when the prompt clearly calls for otherwise. We validate the rubrics directly, reporting 70.8\% rubric-blind judge--human agreement.
\end{itemize}

\paragraph{Outline}
In the rest of this paper, we first review the related work (\cref{sec:related_work}), then describe \bnshort (\cref{sec:benchmark}), followed by a brief section on our experimental setup and evaluation (\cref{sec:setup}) and our results (\cref{sec:results}).
We conclude in \cref{sec:conclusion}.

\section{Related work}
\label{sec:related_work}

The rapid expansion of LLM evaluation into the multimodal domain has been accompanied by a proliferation of benchmarks, recently surveyed in \citet{zhang-etal-2025-redundancy-principles} and \citet{zhang-etal-2025-lmms}.
We organize the most relevant prior work by the modality combinations they target, and then position \bnshort with respect to the small number of existing \emph{cross-modal} benchmarks.

\paragraph{Image-text understanding.}
The bulk of multimodal evaluation effort has gone into vision-language understanding.
General-purpose suites such as MME \citep{fu-etal-2023-mme}, MMBench \citep{liu-etal-2024-mmbench}, MM-Vet \citep{yu-etal-2024-mmvet}, SEED-Bench \citep{li-etal-2024-seedbench}, MMStar \citep{chen-etal-2024-mmstar} and MMMU \citep{yue-etal-2024-mmmu,yue-etal-2025-mmmu} probe a broad set of perception and reasoning skills over (mostly) single-image inputs and short-form text answers.
More targeted benchmarks evaluate specific capabilities, e.g.\ multi-image and visual perception via BLINK \citep{fu-etal-2024-blink}, mathematical reasoning via MathVista \citep{lu-etal-2024-mathvista}, or instruction-following granularity via MIA-Bench \citep{qian-etal-2024-miabench}.
A more recent line of work targets the egocentric, wearable-device setting, where image quality and framing differ substantially from curated third-person benchmarks: WearVQA \citep{chang-etal-2025-wearvqa} evaluates VQA on smart-glasses imagery that may be occluded, poorly lit or blurry, and CRAG-MM \citep{wang-etal-2025-cragmm} extends this to multi-turn, retrieval-augmented question answering over wearable image inputs.
While extensive, all of these benchmarks share a fundamentally bimodal setup: a (potentially complex) visual input is coupled with a textual question and a textual answer.

\paragraph{Video-text understanding.}
Several benchmarks extend the image-text setup to video.
Video-MME \citep{fu-etal-2024-videomme} provides a broad-coverage suite over short, medium and long videos; MVBench \citep{li-etal-2024-mvbench} focuses on temporal reasoning skills; MMBench-Video \citep{fang-etal-2024-mmbenchvideo} extends the MMBench setup to long-form, multi-shot videos; TempCompass \citep{liu-etal-2024-tempcompass} stress-tests fine-grained temporal understanding; EgoSchema \citep{mangalam-etal-2023-egoschema} targets long-form egocentric video question answering; and Perception Test \citep{patraucean-etal-2023-perception} measures perceptual abilities across video and audio.
As with the image-text suites, these benchmarks evaluate text-only outputs in response to a video (and sometimes audio) input.

\paragraph{Audio-text understanding.}
Audio capabilities have historically been evaluated through narrow speech-only suites, but recent work has produced more general audio-language benchmarks.
AudioBench \citep{wang-etal-2025-audiobench}, AIR-Bench \citep{yang-etal-2024-airbench} and MMAU \citep{sakshi-etal-2024-mmau} cover varying combinations of speech, music and environmental sound understanding via instruction-following or QA formats, while Dynamic-{SUPERB} \citep{huang-etal-2024-dynamicsuperb} provides a broad collaborative instruction-tuning benchmark for the speech subset.
Outputs in all of these benchmarks are textual.

\paragraph{Document understanding.}
Document inputs - mixing layout, tables, figures and text - are evaluated by DocVQA \citep{mathew-etal-2021-docvqa}, MP-DocVQA \citep{tito-etal-2023-mpdocvqa}, SlideVQA \citep{tanaka-etal-2023-slidevqa} and the long-document benchmark MMLongBench-Doc \citep{ma-etal-2024-mmlongbench}.
These remain text-output benchmarks over single-document inputs.

\paragraph{Multimodal generation.}
A complementary line of work evaluates models on \emph{producing} non-text outputs, almost exclusively images.
T2I-CompBench \citep{huang-etal-2023-t2icompbench} and GenAI-Bench \citep{li-etal-2024-genaibench} target compositional text-to-visual generation, while MMGenBench \citep{huang-etal-2024-mmgenbench} evaluates LMMs on a text-to-image generation pipeline (asking the model to produce an image-generating caption from a reference image).
With few exceptions, generation benchmarks isolate a single output modality and presuppose that the model is a generator, rather than asking whether a model that \emph{could} generate that modality actually \emph{chooses} to do so when relevant - a question \bnshort directly evaluates (see \cref{sec:benchmark,sec:setup}).

\paragraph{Cross- and any-to-any benchmarks.}
The benchmarks closest in spirit to \bnshort are those that consider multiple input \emph{and} output modalities.
OmniBench \citep{li-etal-2025-omnibench} jointly evaluates audio, image and text understanding but produces text outputs only.
MME-Unify \citep{xie-etal-2025-mme-unify} couples a multimodal understanding suite with a single-modality generation suite (image generation) and a small set of generation-then-understanding ``mixed'' tasks.
MixEval-X \citep{ni-etal-2024-mixeval} aggregates many existing single-modality benchmarks under a unified protocol but does not test cross-modal output decisions within a single prompt.
GIM \citep{patel-etal-2026-gim} similarly reflects a broader move toward harder, more integrative evaluation tasks, but its difficulty comes from coordinating cognitive operations over accessible knowledge rather than from selecting and producing among multiple output modalities.
WorldSense \citep{hong-etal-2025-worldsense} and OmniMMI \citep{wang-etal-2025-omnimmi} target streaming and world-grounded video+audio understanding, again with text outputs.
Outside of academic benchmarks, frontier model release notes \citep{gpt54,claude46,gemini31pro,llama4} typically report only a subset of the bimodal benchmarks listed above and do not report any results that require choosing the correct output modality.

\paragraph{Any-to-any and omni models.}
A parallel line of work has produced models architected for cross-modal input and output, including AnyMAL \citep{moon-etal-2023-anymal}, NExT-GPT \citep{wu-etal-2024-nextgpt}, CoDi \citep{tang-etal-2023-codi}, Macaw-LLM \citep{lyu-etal-2023-macaw}, Unified-IO~2 \citep{lu-etal-2024-unifiedio2} and Chameleon \citep{chameleon2024}.
These models motivate the need for evaluations like \bnshort: as the space of `omni' models grows, there is a corresponding need for benchmarks that probe whether such systems actually choose the appropriate output modality, rather than only asking whether they \emph{can} produce a given modality in isolation.

\paragraph{Tool use and output-format choice.}
A loosely related thread is tool-use evaluation, in which the model must decide \emph{which} tool (or output channel) is appropriate for a given user request, e.g.\ the Berkeley Function Calling Leaderboard \citep{bfcl2024}.
Output-modality selection in \bnshort can be viewed as a multimodal analog: rather than choosing a function call, the model must choose the right combination of asset modalities to deliver in its response.

\paragraph{Positioning of \bnshort.}
\bnshort differs from existing benchmarks along two axes.
First, it spans \emph{five} modalities - text, image, audio, video and document - in both inputs and outputs, with up to three modalities composed within a single prompt; we are not aware of any existing benchmark that covers this combinatorial space.
Second, \bnshort evaluates both whether a model that is marketed as multimodal produces the set of output modalities a self-contained user request calls for, and whether what it returns in each of those modalities actually answers the question.
Human-authored rubric criteria, tagged per output modality, supply the latter judgment, and the \mmiscore averages them across the expected output modalities; the \mps isolates production alone and bounds the \mmiscore from above.
\Cref{tab:benchmark_comparison} summarizes the modality coverage of the academic benchmarks discussed above and contrasts it with \bnshort; \cref{fig:benchmark_coverage} provides the complementary view of benchmarks reported in frontier model release notes.

\begin{table}[t]
\centering
\small
\setlength{\tabcolsep}{4pt}
\resizebox{\textwidth}{!}{\begin{tabular}{l c c c c c c c c c c c}
\toprule
& \multicolumn{5}{c}{\textbf{Inputs}} & \multicolumn{5}{c}{\textbf{Outputs}} & \textbf{Cross-} \\
\cmidrule(lr){2-6}\cmidrule(lr){7-11}
\textbf{Benchmark} & T & I & A & V & D & T & I & A & V & D & \textbf{modal out?} \\
\midrule
MME, MMBench, MMMU \citep{fu-etal-2023-mme,liu-etal-2024-mmbench,yue-etal-2024-mmmu} & \yes & \yes & \no & \no & \no & \yes & \no & \no & \no & \no & \no \\
Video-MME, MVBench \citep{fu-etal-2024-videomme,li-etal-2024-mvbench} & \yes & \no & \kinda & \yes & \no & \yes & \no & \no & \no & \no & \no \\
AudioBench, MMAU \citep{wang-etal-2025-audiobench,sakshi-etal-2024-mmau} & \yes & \no & \yes & \no & \no & \yes & \no & \no & \no & \no & \no \\
DocVQA, MMLongBench-Doc \citep{mathew-etal-2021-docvqa,ma-etal-2024-mmlongbench} & \yes & \no & \no & \no & \yes & \yes & \no & \no & \no & \no & \no \\
T2I-CompBench, GenAI-Bench \citep{huang-etal-2023-t2icompbench,li-etal-2024-genaibench} & \yes & \no & \no & \no & \no & \no & \yes & \no & \no & \no & \no \\
OmniBench \citep{li-etal-2025-omnibench} & \yes & \yes & \yes & \no & \no & \yes & \no & \no & \no & \no & \no \\
MME-Unify \citep{xie-etal-2025-mme-unify} & \yes & \yes & \no & \no & \no & \yes & \yes & \no & \no & \no & \kinda \\
MixEval-X \citep{ni-etal-2024-mixeval} & \yes & \yes & \yes & \yes & \no & \yes & \yes & \yes & \yes & \no & \kinda \\
WorldSense, OmniMMI \citep{hong-etal-2025-worldsense,wang-etal-2025-omnimmi} & \yes & \no & \yes & \yes & \no & \yes & \no & \no & \no & \no & \no \\
\midrule
\bnshort (ours) & \yes & \yes & \yes & \yes & \yes & \yes & \yes & \yes & \yes & \yes & \yes \\
\bottomrule
\end{tabular}
}
\caption{\textbf{Modality coverage of related academic benchmarks.} For each benchmark, we indicate whether it includes Text (T), Image (I), Audio (A), Video (V) and Document (D) on the input and output side, and whether it evaluates the model's choice of \emph{output} modality across multiple non-text options (``Cross-modal out?''). \yes\ indicates support, \no\ indicates no support, and \kinda\ indicates partial support (e.g.\ a separate single-modality generation track that is not jointly evaluated with the understanding side).}
\label{tab:benchmark_comparison}
\end{table}

\paragraph{LLM judges in evaluation.}
Both halves of our scoring pipeline use an LLM judge, and we position that choice against a now-substantial literature.
LLM-as-a-judge has become standard practice in open-ended evaluation since \citet{zheng-etal-2023-mtbench}, with subsequent work documenting both its reliability and its biases \citep[see e.g.][]{li-etal-2024-judgesurvey}.
How much weight the judge carries differs between our two uses of it, and we return to each in \cref{sec:setup}.

\section{\bn: the benchmark}
\label{sec:benchmark}

\begin{table*}[t]
\centering
\small
\renewcommand{\arraystretch}{1.3}
    \begin{tabularx}{\linewidth}{@{}
    X
    >{\raggedright\arraybackslash}m{1.8cm}
    >{\raggedright\arraybackslash}m{2.2cm}
@{}}
\toprule
\textbf{Prompt} & \textbf{Input} & \textbf{Output} \\
\midrule
What's the average salary for my employees?
  & Document & Text \\
Omelet recipe, can you talk me through the steps as I go?
  & Image & Audio \\
Can you help me remove the people from this photo and then turn it into a painting? Please make the sky pink!
  & Text, Image & Image \\
Can you tell me what animal this is and pronounce its name slowly?
  & Audio, Video & Audio \\
I would like to recreate this look. Can you list out all the products that I will need and write out a very detailed instruction in a \texttt{.txt} file?
  & Text, Audio, Video & Document \\
Can you show me how to make soft chocolate chip cookies? A list of ingredients and a tutorial video would be nice.
  & Text & Text, Video \\
My brother sent me this image of his car wheel. Can you help me understand how to read and understand tire sizes? Provide a brief explanation and annotate the picture so I can understand what the size markings on the tire are.
  & Text, Image & Image, Text \\
Explain the science behind convergent plate tectonics, including a 2D diagram and a video simulation of the plates shifting.
  & Text & Image, Text, Video \\
This audio captures a doctor giving important pre-surgery instructions. List some positive things a patient could do before the surgery. Can you also create a video that visually encourages them to stay positive and prepared? Create a printable PDF summary of instructions for patients to refer to later.
  & Audio & Document, Text, Video \\
\bottomrule
\end{tabularx}

\caption{Example prompts from the benchmark, showing the input assets provided to the model and the desired output modalities.}
\label{tab:prompt_examples}
\end{table*}

In total, the \bnshort benchmark contains \nsamples prompts, covering five different modalities, each with a set of human-written rubrics for each desired output modality.
While the largest subset of the collected prompts has one input and one output modality, almost half of the benchmark comprises prompts that have multiple input or output modalities.
All prompts are in English.
In this section, we describe the distribution of prompts over (number of) modalities (\cref{subsec:prompts}), provide more details on the data collection and validation process (\cref{subsec:data_collection}), discuss the rubrics used to assess correctness (\cref{subsec:rubrics}), and define the \mmiscore together with the \mps, the modality-presence diagnostic that bounds it from above (\cref{subsec:scoring}).

\subsection{\bnshort prompts}
\label{subsec:prompts}
\begin{figure}
\begin{subfigure}[b]{0.35\textwidth}
\centering
\includegraphics[width=\textwidth]{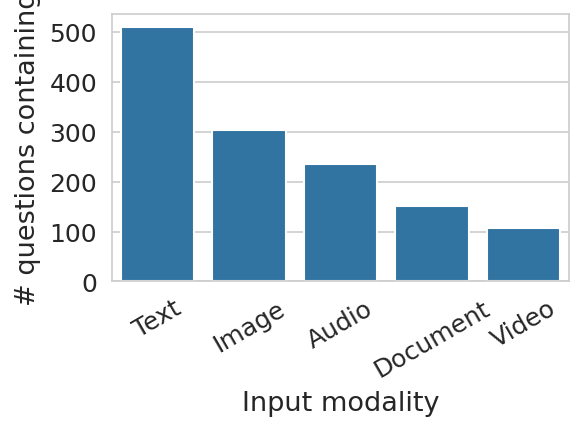}
\caption{}\label{fig:prompts_per_input_modality}
\end{subfigure}
\begin{subfigure}[b]{0.35\textwidth}
\centering
\includegraphics[width=\textwidth]{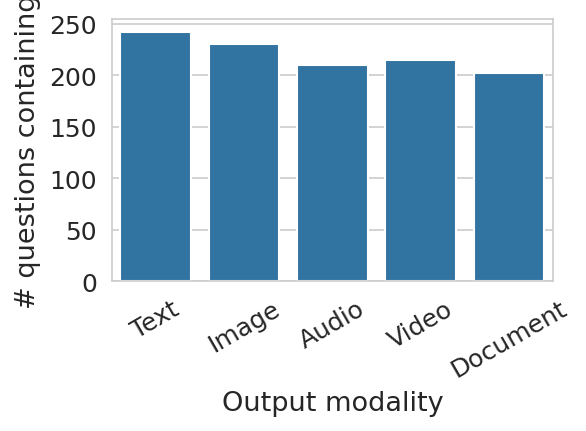}
\caption{}\label{fig:prompts_per_output_modality}
\end{subfigure}
\begin{subfigure}[b]{0.29\textwidth}
\centering
\includegraphics[width=\textwidth]{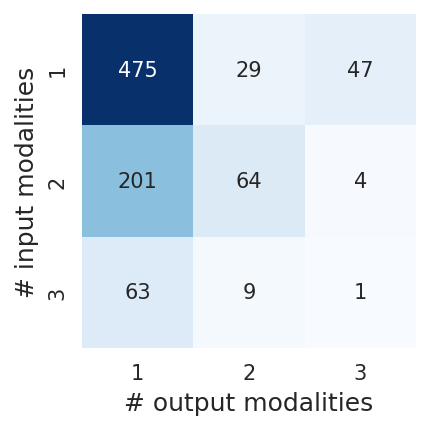}
\caption{}\label{fig:n_modalities}
\end{subfigure}
    \caption{
        \textbf{Benchmark modality statistics.}
        a) Number of times each modality occurs in the input of a \bnshort question.
        b) Number of times each modality occurs in the output of a \bnshort question.
        c) Number of in- and output modalities across \bnshort questions.}
\label{fig:modalities}
\end{figure}

The five different modalities covered in the \bnshort prompts are text, audio, image, video and document\footnote
{The 152 documents in \bnshort cover six file types: \texttt{.pdf} (63 prompts), \texttt{.docx} (48), \texttt{.xlsx} (12),
  \texttt{.pptx} (11), \texttt{.csv} (8) and \texttt{.txt} (10).
}.
As can be seen in \cref{fig:prompts_per_output_modality}, those modalities are more or less evenly covered in the prompts' outputs, with a slight overrepresentation of the text modality, which often accompanies other modalities for clarification or explanation.
In the prompts' inputs, this overrepresentation of text as a modality is even more pronounced (see \cref{fig:prompts_per_input_modality}).
As language is the most natural way for humans to express questions or requests, most prompts with modalities besides text or audio contain text or audio alongside them.

\begin{figure}
\centering
\includegraphics[width=\textwidth]{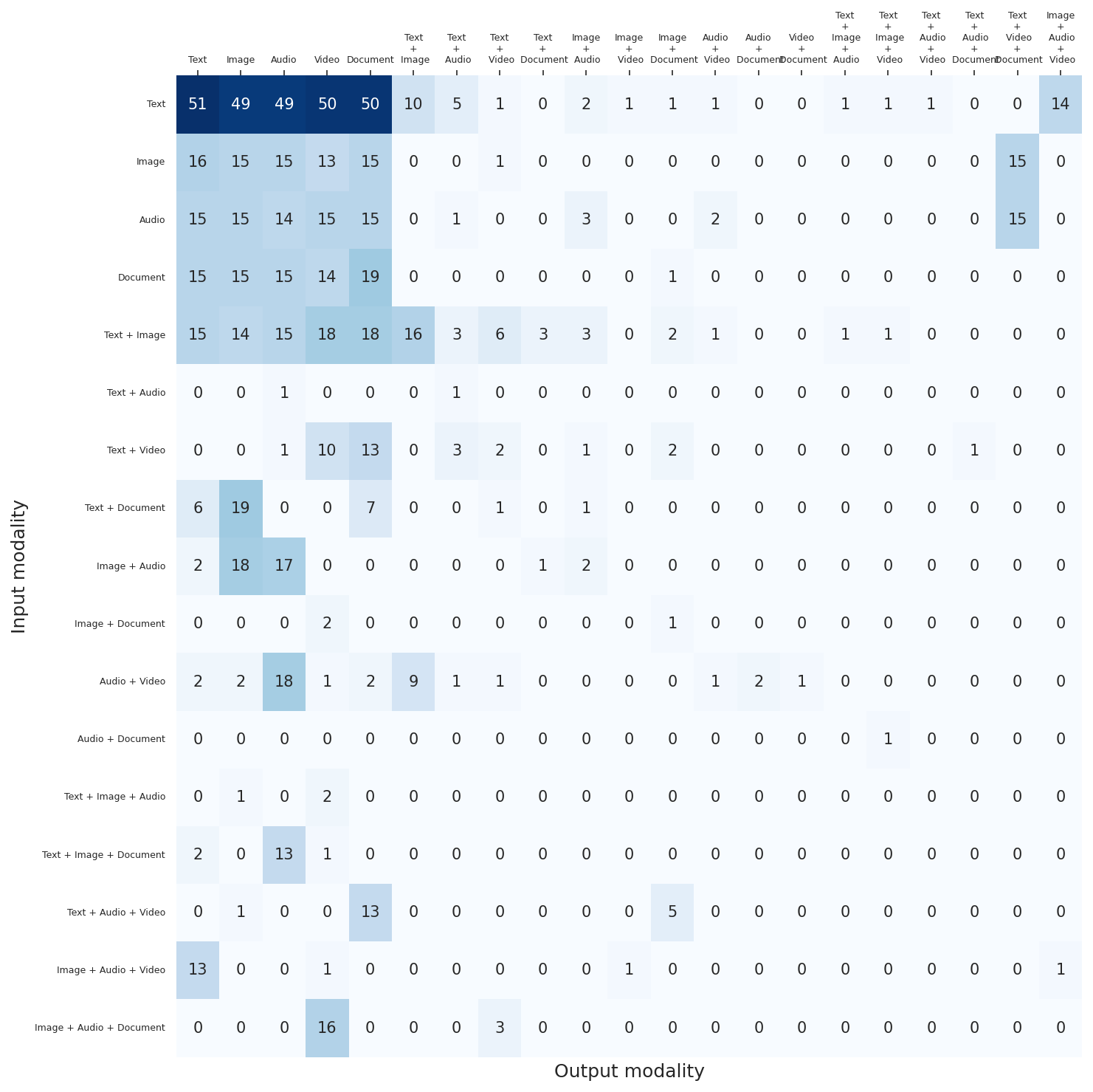}
    \caption{\textbf{Full modality grid.} We show the distribution of input and output modalities of the full \bn benchmark.
    }
\label{fig:modality_heatmap_full}
\end{figure}

In \cref{fig:n_modalities}, we show how many input and output modalities the \bnshort prompts have.
In the top left corner of that figure, we see that the bulk of the prompts have one input and one output modality.
As text remains one of the most common input modalities, 249 of the one-input/one-output prompts have text as input and one of the five modalities as output.
The remaining 226 one-input/one-output prompts are distributed over all the other possible input-output modality combinations.
From the one-input/one-output modality prompts, we gradually increased first the number of input modalities, and then the number of output modalities up to three different modalities.
In total, of the \nsamples prompts, 475 are one-input/one-output, 264 have multiple inputs and a single output, 76 have a single input and multiple outputs, and 78 have multiple inputs and multiple outputs.
A complete picture of coverage across different in- and output modalities can be found in \cref{fig:modality_heatmap_full}.
The prompts themselves can be inspected using the viewer implemented in our \href{https://github.com/facebookresearch/modality-maturity-index}{github repository} or on \href{https://huggingface.co/datasets/facebook/mmi}{HuggingFace}; some representative examples can be found in \cref{tab:prompt_examples}.

\subsection{Data collection}
\label{subsec:data_collection}

We created the data for \bnshort in three stages.
First, we piloted the data collection by creating a small set of 50 questions with text as input and a single modality in the output.
This initial set was created by an internal Meta content team, from the broad instruction to generate questions that would require different modalities to optimally answer them, where the correct output modality should be clear and unambiguous but not explicitly mentioned (e.g.\ avoid questions like `give me a video with...').
Through several iterations of manual inspection by the authors, we guided the content team to create a varied set of 500 questions with one input and one desired output modality.

Using the lessons learned from the first batch of questions, in the next two stages we created questions that have interleaved modalities only in the input, and in both input and output, respectively.
Like in the first stage, we created both these sets starting from a small set of questions to understand potential issues with the collection protocol.
We then manually reviewed and iteratively refined and built upon them to ensure correctness, diversity in question types and coverage across modalities.
Each of the questions in the final set of \nsamples questions has been reviewed manually by at least two of the authors, checking specifically for clarity and correctness of the desired output modalities.\footnote{We did not correct minor issues such as typos that did not affect the clarity of the prompts or expected outputs, as such issues likely also exist in real-world user prompts.}
Disagreements were resolved via discussion.
In the rubric creation round (see \cref{subsec:rubrics}), about 30 gold modalities were adjusted based on input from the annotators writing the rubrics.
The content team consisted of full-time Meta employees compensated as part of their regular employment; all prompts and assets are released.

All assets in the dataset are either generated with MetaAI or created by members of the content team using their phones.
In some cases, we regenerated images or videos taken by the content team with MetaAI. 
While the quality of those images and videos is not always comparable to the originals they replaced, we ensured that they would not make answering the question impossible (e.g.\ because the model is asked to analyze a non-coherent video).

\subsection{Rubrics}
\label{subsec:rubrics}
For each prompt, we also collected a set of rubrics for each of the modalities desired to correctly answer the prompt.
These rubrics allow to assess whether the content of a returned modality is correct, rather than only whether that modality appeared.
They are written to facilitate judging of each output modality in isolation by an LLM judge or human, not requiring access to the question, external knowledge or the other modalities returned by the model.\footnote{For some prompts, fully judging correctness would require assessing several output modalities jointly, for instance checking that generated audio and video are aligned. We do not attempt such judgments, as this would require processing even more combinations of modalities than answering the benchmark prompts themselves.}
We nonetheless supply the judge with the prompt and its input assets for context.
All rubrics are written by an external vendor; the annotation guidelines can be found in \cref{app:rubric_guidelines}.\footnote{The guidelines target 4--12 criteria per output modality. In the released set of 7{,}165 criteria across 1{,}099 (prompt, modality) groups, the median group has 6 criteria and 71.5\% fall in the target band; 10.8\% have fewer than three and 6.0\% more than twelve. We did not enforce the target during collection.}
In \cref{subsec:rubric_evaluation}, we report judge--human agreement between judges using the rubrics and humans annotating outputs without a rubric.

\begin{tcolorbox}[
    enhanced,
    colback=gray!5,
    colframe=black!70,
    colbacktitle=black!70,
    coltitle=white,
    fonttitle=\bfseries,
    title={Example rubric set (desired output modalities: video, image, audio)},
    breakable,
    sharp corners=downhill,
    boxrule=0.5pt,
    left=6pt, right=6pt, top=6pt, bottom=6pt,
]
\small
\textit{``I'm learning to crochet. Generate a clip of different patterns to use and how to manuever the fingers. Have some still images of the patterns for reference. Also, how do you pronounce crochet?''}
\begin{enumerate}[nosep,leftmargin=1.8em,topsep=4pt]
    \item \textit{(video)} Provides a video clip for learning crochet patterns and finger movements.
    \item \textit{(video)} Shows multiple crochet patterns in the video.
    \item \textit{(video)} Demonstrates how to maneuver the fingers, yarn, and hook during crocheting in the video.
    \item \textit{(image)} Provides images of the crochet patterns.
    \item \textit{(audio)} Pronounces the word crochet correctly as kroh-SHAY.
\end{enumerate}
\end{tcolorbox}

\subsection{Scoring MMI}
\label{subsec:scoring}
The \mmiscore is main score of the benchmark: it quantifies the extent to which a model answered the question correctly.
The \mps instead focuses only on modality presence, measuring only whether the correct set of output modalities is returned rather than if their content is correct.
Both are implemented in our github repository. 
Below, we discuss them in more detail.

\subsubsection{The \mmiscore}
\label{subsec:mmi_score}
The \mmiscore quantifies overall model correctness, assigning equal weight to each modality.
To circumvent the need for human annotators, we compute \mmiscore using an LLM judge, who grades the answers based on a set of modality-specific rubric criteria (\cref{subsec:rubrics}), 
A criterion is graded in isolation, meaning the judge sees the model's output for that one modality and nothing it produced for the others, and returns a score in $[0,1]$.
A prompt's \mmiscore is macro average of the scores for each desired output modality; a model's \mmiscore is the average of those per-prompt values.
If a desired modality is absent in a model's output, it's score is set to 0.
This makes the \mmiscore little informative for models that return very few assets: a low score does not tell us whether the answers for returned modalities are incorrect or simply missing.
We therefore complement it with a second score that isolates the missing-modality component.

\subsubsection{The \mps and complementary metrics}
\label{subsec:mps}
The \mps quantifies only how often the correct modalities are returned, ignoring entirely whether they are correct or not.\footnote{
How we determine which modalities a response contains is described in \cref{subsec:modality_detection}.
}
We define \mps as macro-average of the F1 between the set of modalities a model returns for an \bnshort prompt and the gold set of modalities required for the prompt.
For completeness, we also report precision, recall and a separate \emph{pass rate} which indicates whether all expected modalities are present in the model response.\footnote{Unlike recall, pass rate is a binary per-prompt metric: it is 1 if and only if all desired modalities are present, and 0 otherwise.}
Where recall tells us how often the desired modalities are included in the model's response, precision indicates how often the modalities returned by a model are in fact desired.
In computing precision, we do not penalize undesired text, so points for text are essentially `free'.
Some models cannot process some of the input modalities in the prompt set (\cref{tab:supported_modalities}); the provider rejects the call and the prompt is scored zero, because being unable to ingest a modality is itself a deficiency in handling it.
We report how often this happens as the \emph{input failure rate}, in the last column of \cref{tab:overall_metrics}.


\section{Experimental setup}
\label{sec:setup}

We run two experiments.
First, in our main evaluation (\cref{subsec:models,subsec:modality_detection}), we apply \bnshort to five models marketed as multimodal, under the benchmark's normal protocol, and report \mps scores.
As this experiment returns very few gradeable assets, we run a second experiment, with the sole purpose of validatiing the rubrics (\cref{subsec:rubric_eval_setup}).
This \emph{rubric-validation run} is a separate experiment on three different models given generation tools; its purpose is to produce enough gradeable assets to put the rubrics to the test, and it is not a model comparison.\footnote{Given the different purposes of these experiments and the different settings, the numbers of these experiments are thus not comparable.}

\subsection{Main evaluation: models and generation settings}
\label{subsec:models}
The five models we evaluated are GPT-5.4 \citep{gpt54}, Claude Opus 4.6 \citep{claude46}, Gemini 3.1 Pro \citep{gemini31pro}, Gemini 3.1 Flash \citep{gemini31flash} and Llama 4 Maverick \citep{llama4}.
The first four were run through their respective provider APIs.
Llama 4 Maverick is open-weight and has no first-party API of the same kind, so we served the \texttt{Llama-4-Maverick-17B-128E-Instruct-FP8} checkpoint ourselves on local hardware and queried that deployment instead.
All models were run with the model's default parameters, except for the generation limit, which we set to 32{,}768 tokens throughout.\footnote{Provider calls use retry with transient-error keyword matching (e.g.\ rate limit, 5xx, timeout-related signals), with a maximum of five retries, and a runner-level request time-out of 300 seconds per call. If retries are exhausted or a non-retryable exception occurs, the prompt is recorded as a failure and counts toward the input failure rate of \cref{tab:overall_metrics}.}

\subsection{Detecting modalities in model outputs}
\label{subsec:modality_detection}
We have three modes of detecting whether a model outputs a particular modality.
First, we look directly at the assets a model returns.
When a model responds with an asset of the correct modality, we mark that modality as \emph{natively} generated.
Sometimes, models return links to assets of the right modality, which we detect using regexes for platforms that host assets of that specific modality, such as YouTube for videos or Flickr for images.
If nothing is detected using either of these two options, we pass the answer to Gemini 3 Flash and ask it to judge whether an asset of a particular modality was generated.\footnote{Model ID \texttt{gemini-3-flash-preview}. The same judge model is used for the rubric-validation run in \cref{subsec:rubric_eval_setup}.}
The full prompt can be found in \cref{app:judge_prompt}.

These three routes give us two standards for counting a modality as present.
Under the \texttt{strict} standard, only natively returned assets count: a modality is present if and only if the model itself produced an asset of that type.
Under the \texttt{lenient} standard, a modality counts as present if \emph{any} of the three routes fires: a returned asset, a link to an asset of that modality, or a fallback judge detection.
Unless stated otherwise, every presence number we report is lenient, including the \mps values in \cref{tab:overall_metrics} and \cref{fig:overall_rpf}; the sole exception is \cref{fig:heatmap-precision-native}, which reports strict (native) precision.
We take lenient as the default because it reflects what a user actually receives: a working link to a video of the right content answers the request, whether or not the model rendered the pixels itself.
Note that this also implies that even models that do not have the ability to natively generate specific modalities (e.g.\ video) can still answer questions that require that modality as output correctly by responding with a link to a pre-existing asset.
Whereas for some questions this may not be appropriate (e.g.\ if the prompt asks to edit a video), for others it is instead the desired mode of responding (e.g.\ when a prompt asks for a video of an interview with an existing person).

\begin{table}
    \centering
    \begin{tabular}{l *{2}{c} *{2}{c} *{2}{c} *{2}{c} *{2}{c}}
\toprule
& \multicolumn{2}{c}{Text}
& \multicolumn{2}{c}{Image}
& \multicolumn{2}{c}{Audio}
& \multicolumn{2}{c}{Video}
& \multicolumn{2}{c}{Document} \\
\cmidrule(lr){2-3} \cmidrule(lr){4-5} \cmidrule(lr){6-7} \cmidrule(lr){8-9} \cmidrule(lr){10-11}
Model & In & Out & In & Out & In & Out & In & Out & In & Out \\
\midrule
GPT-5.4           & \yes & \yes & \yes & \yes & \no  & \no & \no  & \no & \kinda & \no  \\
Gemini 3.1        & \yes & \yes & \yes & \no & \yes & \no & \yes & \no  & \yes  & \no  \\
Claude Opus 4.6   & \yes & \yes & \yes & \no & \no  & \no  & \no  & \no & \kinda & \no  \\
Llama 4 Maverick  & \yes & \yes & \yes & \no & \no  & \no  & \yes & \no  & \kinda  & \no  \\
\bottomrule
\end{tabular}

    \caption{\textbf{Natively supported modalities.} Modalities natively supported by the different models. \yes\ indicates the model supports the modality, \no\ indicates it does not; \kinda indicates that the modality is partly supported, e.g.\ a model accepts \texttt{.txt} and \texttt{.pdf} documents, but not \texttt{.docx}.}
    \label{tab:supported_modalities}
\end{table}

\subsection{Rubric-validation run: tool scaffolding}
\label{subsec:rubric_eval_setup}
Because the models in our main evaluation produce so few assets, their answers are not sufficient to judge if our rubrics grade well.
We therefore set up an additional experiment in which we give three models access to tools that help them generate audio, image and video, respectively.
The purpose is not to improve model performance or to rank models, but to increase coverage across output modalities so that the rubrics can be exercised.
The models that we use for this experiment are Claude Opus 4.8, Gemini 3.5 Flash and GPT-5.4.
Each of them is given three tools, named \texttt{image\_gen}, \texttt{audio\_gen} and \texttt{video\_gen}, that are exposed as generic function calls: each takes a single free-text \texttt{prompt} argument and carries an identical, uninformative description, so that the only signal about a tool is its name and the modality that name implies.
Behind that interface we route the prompt to a backend generator (Gemini 3.1 Flash Image for images, Gemini 2.5 Flash TTS for audio and Veo 3.1 Fast for video), capture the returned asset, and hand it back to the calling model.
Tool use runs for at most three rounds per prompt: in each round we dispatch every tool call the model requested, return the resulting assets, and let it continue, stopping early once it requests no further tools.
A round may contain several calls, so a prompt can yield more than three generated assets; in practice the models requested tools on 189--502 of the \nsamples prompts and issued a median of one to three calls when they did.
The calling model is never told which backend it is addressing, what that backend is good at, or how to prompt it, limiting any benefit from recognizing a familiar tool API or from backend-specific instructions in the tool description.
We use the same generation settings as in the main evaluation (32{,}768-token limit, up to 5 retries), with an extended 900-second request time-out to accommodate slower video and audio generation.
We then collect all assets and text and ask both human annotators and an LLM judge (Gemini 3 Flash) to provide per-modality judgments on the correctness of the responses.
The LLM judge uses the human-written rubrics, grading one criterion at a time with the prompt given in \cref{app:rubric_judge_prompt}, while the humans judge the answers without rubric (annotation instructions are provided in \cref{app:annotation_guidelines}).
We use the resulting agreement between humans and judges to quantify how well LLM judges can approximate \mmiscore given the rubrics.

\section{Results}
\label{sec:results}

Following the division set out in \cref{sec:setup}, \cref{subsec:overall_presence,subsec:input_failures,subsec:output_modality} report the main evaluation of five models and \cref{subsec:rubric_evaluation} the rubric-validation run.
In \cref{tab:overall_metrics}, we report a summary table with all aggregate metrics we computed for the five main-evaluation models, sorted by \mps (per-prompt mean F1).
We will discuss these metrics as well as further breakdowns along specific axes in the subsequent sections.

\begin{table}
\centering
\begin{tabular}{l c c c c c}
\toprule
Model & F1 & Recall & Precision & Pass rate & Input failure rate \\
\midrule
GPT-5.4 & 34.9 & 33.8 & 99.8 & 31.0 & 30.1\% \\
Gemini 3.1 Flash & 27.2 & 25.4 & 99.4 & 20.3 & 8.2\% \\
Gemini 3.1 Pro & 25.4 & 23.6 & 99.7 & 18.6 & 8.0\% \\
Llama 4 Maverick & 16.3 & 15.1 & 100 & 12.0 & 36.5\% \\
Claude Opus 4.6 & 15.6 & 14.5 & 100 & 11.3 & 37.0\% \\
\bottomrule
\end{tabular}

    \caption{\textbf{Summary model scores.} Per-model sample-level averages of F1 (the \mps), recall, precision, pass rate and input failure rate. Models are sorted by \mps.}
\label{tab:overall_metrics}
\end{table}

\subsection{Overall presence scores}
\label{subsec:overall_presence}
In \cref{fig:overall_rpf}, we show the overall \mps (F1), precision and recall for all models.
The precision of all models is at or near 100, indicating that, apart from text, models rarely generate any modality that was not in the list of expected modalities.
The models' recall is substantially lower: GPT-5.4, Gemini 3.1 Flash, Gemini 3.1 Pro, Llama 4 Maverick and Claude Opus 4.6 obtain recall scores of 33.8, 25.4, 23.6, 15.1 and 14.5 respectively, with no model exceeding 35.
Because precision is this high, the \mps stays within two points of recall for every model, so on this set of models the two support the same conclusions.

\begin{figure}[t]
    \centering
    \begin{subfigure}[b]{0.30\textwidth}
        \includegraphics[height=4.0cm, trim=0 0 47 0, clip]{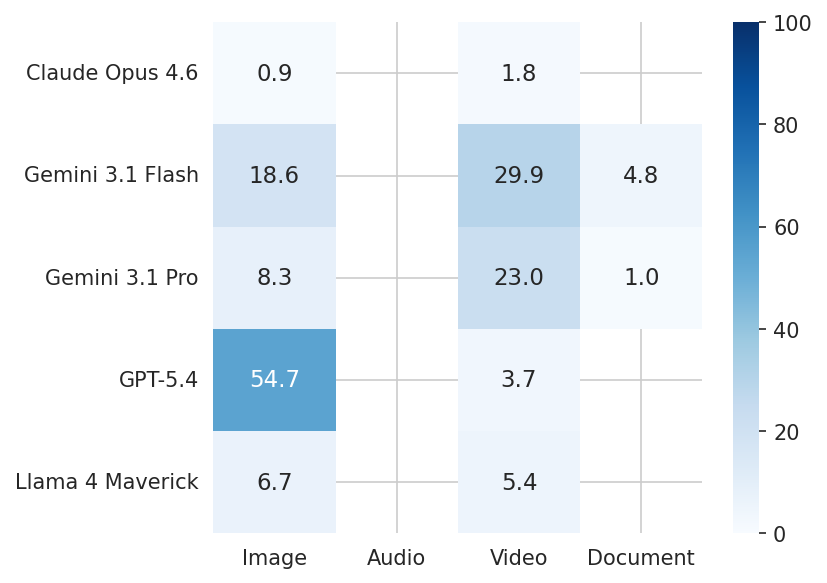}
        \caption{F1}
        \label{fig:heatmap-f1}
    \end{subfigure}
    \hfill
    \begin{subfigure}[b]{0.21\textwidth}
        \includegraphics[height=4.0cm, trim=100 0 47 0, clip]{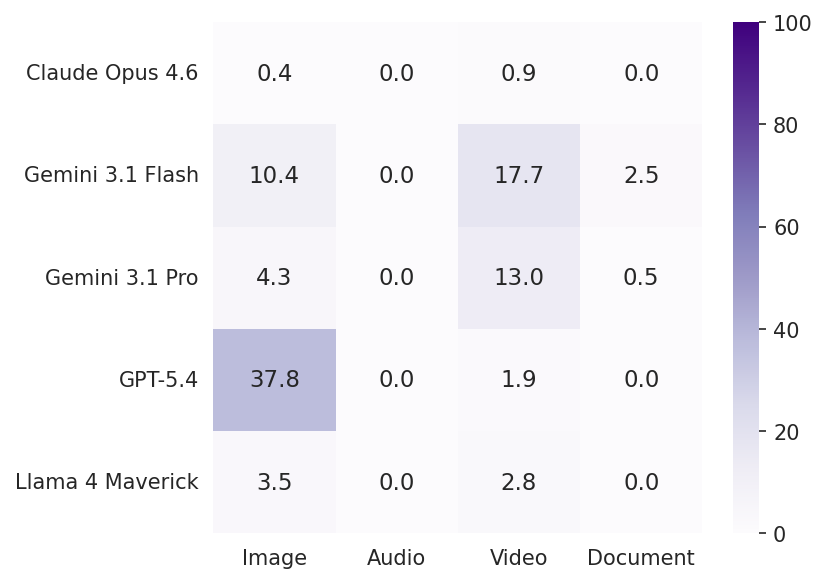}
        \caption{Recall}
        \label{fig:heatmap-recall}
    \end{subfigure}
    \hfill
    \begin{subfigure}[b]{0.21\textwidth}
        \includegraphics[height=4.0cm, trim=100 0 47 0, clip]{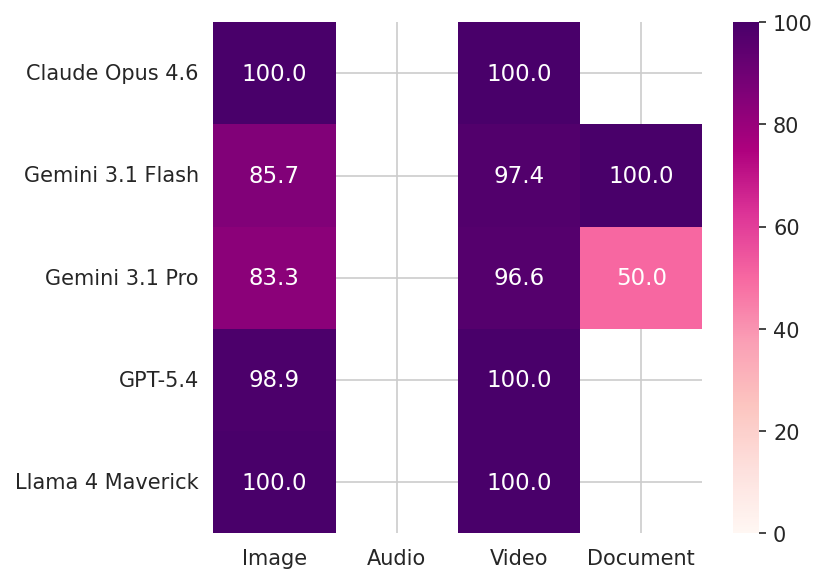}
        \caption{Precision}
        \label{fig:heatmap-precision}
    \end{subfigure}
    \hfill
    \begin{subfigure}[b]{0.21\textwidth}
        \includegraphics[height=4.0cm, trim=100 0 47 0, clip]{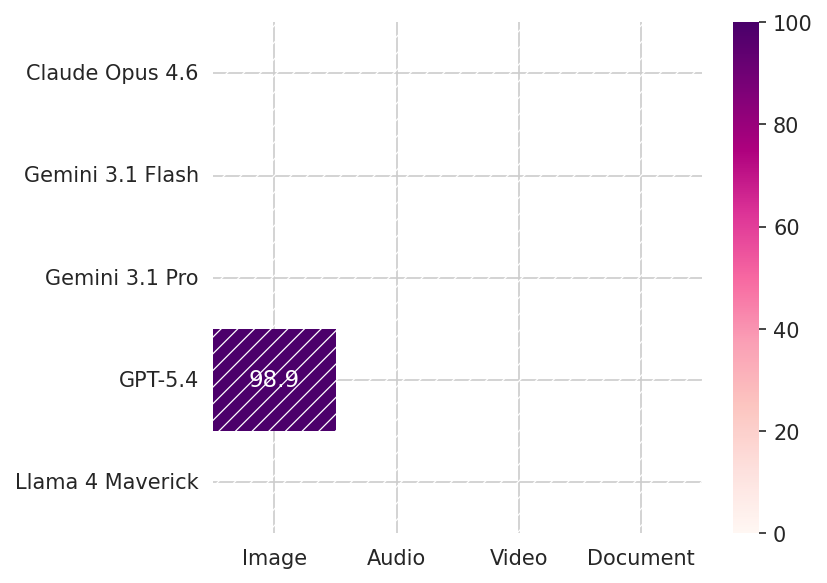}
        \caption{Precision native}
        \label{fig:heatmap-precision-native}
    \end{subfigure}
    \caption{\textbf{Performance by model and output modality.} We show F1, recall and precision per model and output modality. Apart from text, GPT-5.4 is the only evaluated model to natively generate another desired modality (images), though the models do sometimes return links to images and videos.}
    \label{fig:per-modality-heatmaps}
\end{figure}

\subsection{Input failure rate and pass rate}
\label{subsec:input_failures}
The input failure rates in \cref{tab:overall_metrics} vary substantially across models, from 8.0\% for Gemini 3.1 Pro to 37.0\% for Claude Opus 4.6.
Read against recall, the input failure rate separates two distinct deficiencies: a model that cannot accept the input at all, and one that accepts it and then returns the wrong output modality.

Pass rate provides the complementary prompt-level view, asking whether a response contains \emph{every} expected output modality rather than averaging success across them.
It falls from 31.0 for GPT-5.4 to 11.3 for Claude Opus 4.6 and is consistently below recall, showing that partial fulfillment is common: models may return one expected output modality while omitting another.
Like recall, pass rate counts an input failure as a failure, so it does not by itself separate requests a model cannot accept from those it accepts but only partly fulfils; that separation comes from reading it against the input failure rate.

\subsection{The impact of output modality}
\label{subsec:output_modality}
\begin{figure}[t]
    \centering
    \includegraphics[width=\linewidth]{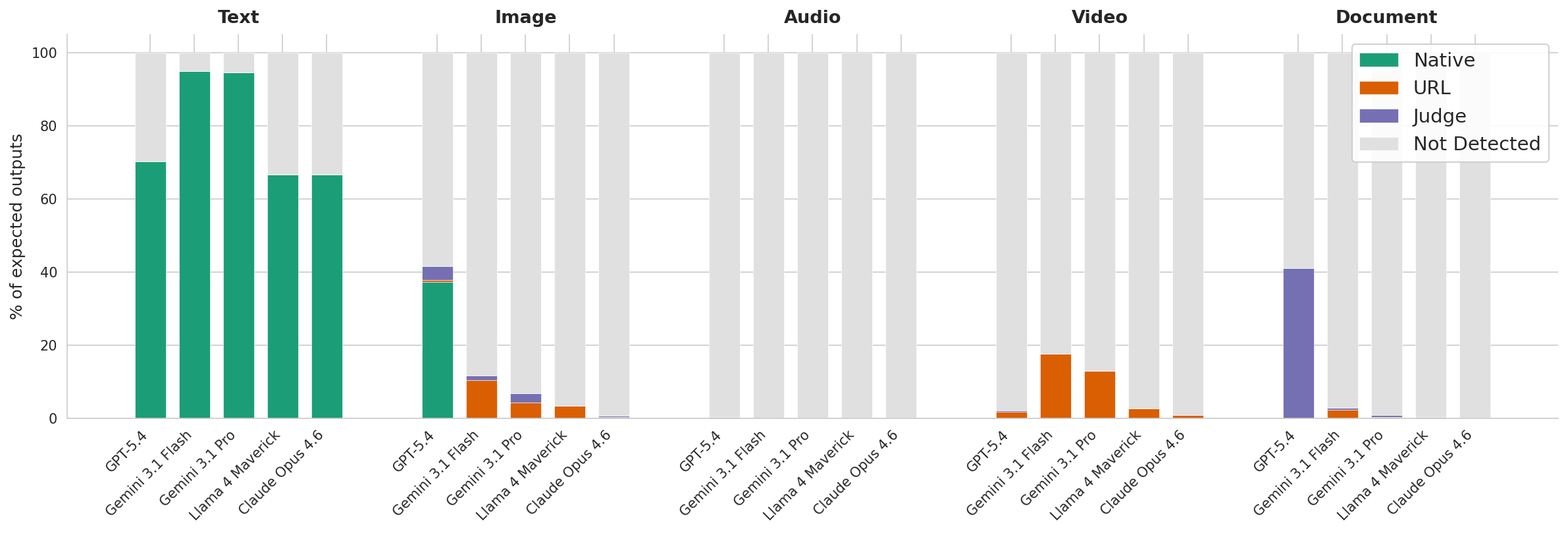}
    \caption{\textbf{Detection method per modality.} Models can return assets of the right modality natively or via a URL. As a fallback, we ask a judge to parse the answer and detect whether an asset of the correct modality is returned.}
    \label{fig:detection-method-per-modality}
\end{figure}

In \cref{fig:per-modality-heatmaps}, we show F1, recall, precision and native precision per output modality.
To compute these values, we average across the entire dataset for each modality, rather than averaging per-prompt scores as for the main \mps.
E.g.\ video recall indicates the percentage of times a model returns a video when a video was a desired output modality, whereas video precision indicates the percentage of times a returned video was in fact in the correct set of output modalities.
When a model never returns a specific output modality, we set precision to NaN rather than 0; NaN cells are excluded from any cross-modality averages.
As none of the models supports all output modalities natively, recall is expectedly low across the board.
Notably, however, not all F1 and recall values are zero for unsupported output modalities.
As can be seen in \cref{fig:detection-method-per-modality}, image and video credit comes from models returning links to existing assets, whereas GPT-5.4's document credit comes almost entirely from the fallback judge recovering generated files that asset- and URL-based detection missed.

Because so few modalities are natively supported, it is hard to say which modalities are more difficult.
As can be seen in \cref{fig:heatmap-precision-native}, GPT-5.4 is the only model that ever natively generates any modality besides text.
The per-modality F1, precision and recall therefore mostly reflect the (often small) fraction of cases where a model is able to return the modality at all, rather than a true measure of its difficulty.
Audio, not supported as an output modality by any of the models, is the most difficult modality across the board: no model generates even links to audio snippets.
In sum, the plots show clearly that there is much room for improvement for any non-text modality, exemplifying the need for more benchmarks such as \bnshort.

\subsection{Rubric validation results}
\label{subsec:rubric_evaluation}
The main-evaluation models produce too few assets for their \mmiscores to be worth grading (\cref{subsec:mmi_score}), so we validate the rubrics on the tool-scaffolded run of \cref{subsec:rubric_eval_setup} instead.
The scaffolded models produced an output for 71\% of the modalities, or 76\% once documents, for which we provided no tool and which are never produced, are set aside, against about a quarter of expected modalities in the main evaluation.
For completeness, we record the \mmiscores these three runs obtain in \cref{app:mmi_scores}.
We compare the two graders on the 1{,}499 (prompt, model, modality) triples for which both returned a judgment and the model actually produced the modality. 
The two graders do not natively produce the same quantity: the humans produce a per-modality binary judgement (see \cref{app:annotation_guidelines} for their annotation guidelines), the judges provide judgements for each rubric criterion, resulting in graded judgements for each modality.
To align the two types of judgements, we collapse the judge grades to a binary judgement, assigning a score of 1.\ if all rubric criteria for a modality are graded 1.\ and 0.\ otherwise.\footnote{
This is the strictest choice possible, but most aligned with the task of the humans.
Empirically, it does not appear to influence the number negatively in a significant way: a sweep over the collapse threshold resulted in a maximum agreement of 73.2\% for $t=0.8$ and Scott's $\pi$ at 0.43 for $t=0.9$.}
Because the judge is itself a Gemini model and grades a plurality of assets from its own family, we cannot rule out a self-preference effect \citep[see e.g.][]{li-etal-2024-judgesurvey}; the agreement figures should therefore not be read as judge-independent.

\begin{figure*}[t]
\centering
\begin{subfigure}[c]{0.48\linewidth}
    \centering
    \begin{tabular}{l r r r}
\toprule
Modality & N & \% Agree & $\pi$ \\
\midrule
Text & 519 & 70.9 & 0.34 \\
Image & 415 & 66.5 & 0.28 \\
Audio & 284 & 73.6 & 0.45 \\
Video & 281 & 74.4 & 0.31 \\
\midrule
Overall & 1499 & 70.8 & 0.41 \\
\bottomrule
\end{tabular}
    \caption{}\label{tab:judge-human}
  \end{subfigure}
  \begin{subfigure}[c]{0.35\linewidth}
    \centering
    \includegraphics[width=\linewidth]{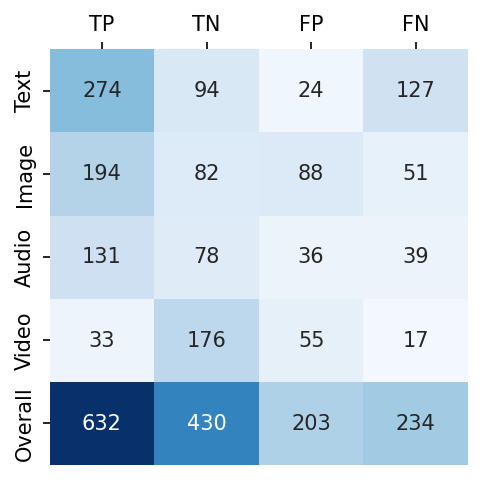}
    \caption{}\label{fig:judge-human-confusion}
  \end{subfigure}
\caption{\textbf{Judge--human agreement on \bnshort correctness}
    a) Per modality judge--human agreement (\% agreement and Scott's $\pi$) between the LLM judge and human annotators.
    b) Per modality breakdown of true and false positives and negatives.
    Document is absent from both panels: we provided no document-generation tool, so no model in this experiment produced a document output and the document rubrics could not be exercised (\cref{subsec:rubric_evaluation}).
    }
\label{fig:judge-human}
\end{figure*}

Following \citet{thakur-etal-2025-judging}, we compute both percent agreement and Scott's $\pi$ between the human and LLM judgements.
In \cref{tab:judge-human}, we can see that the judge and the human agree on 70.8\% of judgments overall, ranging from 66.5\% on image to 74.4\% on video.
Scott's $\pi$ is 0.41 overall (last column of \cref{tab:judge-human}), lower because it normalizes for chance agreement: as can be inferred from the last row of \cref{fig:judge-human-confusion}, both the LLM judge and the humans mark between 50 and 60\% of outputs correct, putting their chance agreement near 50\%.
As document is missing entirely from the generated assets, the document rubrics unfortunately remain unvalidated.


We conduct a small analysis of the differences between human and LLM judgments, looking at 20 samples in which the human assessment was `correct' and the judge `incorrect', and vice versa.
In the former category, cases in which the LLM marks a sample incorrect while the human marks it correct, a few cases are LLM judgment errors (e.g.\ stating that a piece of text has three paragraphs while in fact it has two plus a title), and a few involve prompts where correctness is somewhat subjective.
The rest, rather than being real errors, mostly illustrate the difficulty of writing rubrics for complex requests that precisely describe when an answer is correct, especially when there are multiple different ways of correctly answering the request.
The latter category, cases in which the LLM marks a sample correct while the human marks it incorrect, paints a clearer picture.
Here too, some cases are judge mistakes (e.g.\ stating that an audio file is 20 seconds while it is in fact 30).
However, most involve prompts that request an edit or adaptation of an input image or video, where the response does not adhere to the original image or video.
The judge does receive the prompt's input assets, so this is not a matter of missing context.


\section{Conclusion}
\label{sec:conclusion}

Prompted by the lack of cross-modal benchmarks for LLMs, in this paper we propose the \benchmarkname\ (\bnshort), a multimodal benchmark with \nsamples prompts spanning up to three out of five possible in- and output modalities.
The \bnshort prompts aim to comprise a diverse set of possible user questions, covering different topics and modality combinations in input as well as output.

Our main evaluation reports the \mps, a score quantifying how often models return the correct set of modalities.
This metric shows that even without considering the \emph{correctness} of the answers, \bnshort is challenging for most models, with the \mps (per-prompt mean F1) ranging from 15.6 (Claude Opus 4.6) to 34.9 (GPT-5.4).
We analyze the difficulty of various output modalities and study the difference between model precision and recall.
We find that low scores are mostly recall driven: many modalities are simply not supported by some of the models, and even if they are supported, they are not always used correctly.
The most challenging modality is audio, which none of the evaluated models supports as an output and which is not returned for any prompt that requires it.
Image and video receive higher scores, but are still very challenging on the output side even for the highest scoring model.

Because these models return so few assets, we validate the rubrics behind the \mmiscore in a separate experiment.
We find that an LLM judge using the rubrics agrees with a human annotator on 70.8\% of judgments (Scott's $\pi$ of 0.41), ranging from 66.5\% on image to 74.4\% on video.
Our analysis indicates that the remaining gap comes from judge errors, from the difficulty of specifying correctness for open-ended requests, and from the difference between a rubric-driven judgment of a single modality and an unaided human one.
We expect agreement to go up as judges get better.

The results on \bnshort paint a stark picture: despite being marketed as multimodal or even `omni' models, current LLMs remain predominantly text-output in practice.
The gap between what is advertised and what is delivered is substantial, not because models confuse modalities, but because they fail to respond with non-text modalities even when the question clearly calls for them.
As multimodal interactions become more common in user-facing applications, closing this gap is a natural target for future model development.
\bnshort is created to aid in making such progress and is, beyond providing a snapshot of current capabilities, designed to serve as a living benchmark for tracking multimodal progress over time.

\subsection*{Limitations}
Creating an interleaved multimodal benchmark and appropriately evaluating it comes with many challenges.
While we have tried to address those to the best of our abilities, our work knows limitations as well.

\begin{itemize}[leftmargin=*,topsep=2pt,itemsep=2pt]
    \item \textbf{The benchmark's main index is defined but not evaluated.} The main numbers we report in our work are \mps values, which express only whether the expected modalities were returned. The evaluated models return too few assets for an \mmiscore to be meaningful (\cref{subsec:mmi_score}), and while this is a limitations of models perhaps more than a limitation of the benchmark, it does imply that the actual main score of the benchmark is not well tested. We assessed the rubrics through human-judge agreement, but the agreement was far from perfect, leaving some open questions regarding the viability of computing the \mmiscore automatically using juges.
    \item \textbf{We evaluate APIs, not products.} All five models we tested are queried through their provider API with bare settings, no system prompt and no tools. While this is where also their claim to omni-modality is made, deployed assistants often wrap these models with image, audio or video generators, which would likely largely increase the \mps as well as \mmiscore. The scores therefore characterize the model as exposed by the API, not the assistant a user interacts with.
    \item \textbf{Answer correctness may subjective.} Some prompts have an unambiguous answer, but others are open-ended, and reasonable users may disagree about what a correct answer should involve and -- while we tried to avoid this as much as possible -- even about what modalities may be acceptable to do so. This adds a degree of subjectivity to the benchmark which is not properly evaluated given the low \mps of the models.
\end{itemize}

\subsection*{Future work}
Most of the future work we envision is contigent on better models: as \mps scores go up, this will likely allow us to improve both the benchmark as well as the scoring.
One potential immediate way to do that would be to evaluate tool-augmented and agentic configurations directly, evaluating complete products more than the models underpinning it.
Such models could be evaluated as judges as well as on the benchmark.
A new generation of models may also warrant revising the rubrics themselves.
Another important future direction involves creating benchmark data beyond English.

\subsection*{Acknowledgments}
We thank \textbf{Di Lin}, \textbf{Kieran Nash}, \textbf{Tara Fowler}, and \textbf{Manohar Paluri} for their support.

\clearpage
\newpage

\clearpage
\newpage
\beginappendix

\section{Modality Detection Judge Prompt}
\label{app:judge_prompt}

Below, we show both the judge system prompt and judge user prompt template passed to Gemini 3 Flash (\texttt{gemini-3-flash-preview}) to judge whether a model returned an asset of the correct modality.

\begin{tcolorbox}[
    enhanced,
    colback=gray!5,
    colframe=black!70,
    colbacktitle=black!70,
    coltitle=white,
    fonttitle=\bfseries,
    title=Judge System Prompt,
    breakable,
    sharp corners=downhill,
    boxrule=0.5pt,
    left=6pt, right=6pt, top=6pt, bottom=6pt,
]
\ttfamily\small
You are a strict evaluator for a multimodal AI benchmark. Your job is to examine a model's API response and determine which output modalities it genuinely produced.

\medskip
The five modalities are: Text, Image, Audio, Video, Document.

\medskip
For each modality, look for concrete evidence the model actually produced it:
\begin{itemize}[leftmargin=*,nosep]
    \item \textbf{Text:} Any substantive text output (not just a refusal or error message).
    \item \textbf{Image:} Native image data, base64-encoded image content, complete SVG markup, or a URL pointing to a generated image.
    \item \textbf{Audio:} Native audio data, base64-encoded audio content, or a URL pointing to generated audio.
    \item \textbf{Video:} Native video data, base64-encoded video content, or a URL pointing to generated video.
    \item \textbf{Document:} A generated file such as PDF, CSV, HTML, DOCX, or structured data clearly intended as a downloadable document.
\end{itemize}

\medskip
Do NOT count a modality if the model only wrote ABOUT it. The model must have produced actual media data or file content. Describing what an image would look like, writing a screenplay for a video, or providing lyrics instead of audio are NOT productions of those modalities.

\medskip
A refusal (``I can't generate images'') is NOT a production.

\medskip
Respond with ONLY valid JSON:
\begin{center}
\texttt{\{"detected\_modalities": [], "reasoning": "brief explanation"\}}
\end{center}

The \texttt{detected\_modalities} array should list every modality the model genuinely produced. An empty array means the model produced nothing usable. Valid modality names: Text, Image, Audio, Video, Document.
\end{tcolorbox}

\begin{tcolorbox}[
    enhanced,
    colback=gray!5,
    colframe=black!70,
    colbacktitle=black!70,
    coltitle=white,
    fonttitle=\bfseries,
    title=Judge User Prompt Template,
    breakable,
    sharp corners=downhill,
    boxrule=0.5pt,
    left=6pt, right=6pt, top=6pt, bottom=6pt,
]
\ttfamily\small
The model was asked:\\
``\{prompt\_text\}''

\medskip
The model's response text was:\\
\rule{\linewidth}{0.4pt}\\
\{text\_content\}\\
\rule{\linewidth}{0.4pt}

\medskip
The model's structured/raw API response (JSON, truncated) was:\\
\rule{\linewidth}{0.4pt}\\
\{raw\_response\_json\}\\
\rule{\linewidth}{0.4pt}

\medskip
Which of the five modalities (Text, Image, Audio, Video, Document) did the model genuinely produce in this response?
\end{tcolorbox}

\section{Rubric Judge Prompt}
\label{app:rubric_judge_prompt}

Below, we show the system prompt and user prompt template passed to Gemini 3 Flash (\texttt{gemini-3-flash-preview}) when grading a model response against the human-written rubrics of \cref{app:rubric_guidelines}.
The judge is called once per rubric criterion: each call carries a single criterion and only the model-produced payload for the modality that criterion is tagged with, never the model's output for the other modalities.
For \emph{text} criteria the payload is the response text itself; for the other modalities the generated artifact is attached to the call as media, alongside the prompt's own input assets as context.

\begin{tcolorbox}[
    enhanced,
    colback=gray!5,
    colframe=black!70,
    colbacktitle=black!70,
    coltitle=white,
    fonttitle=\bfseries,
    title=Rubric Judge System Prompt,
    breakable,
    sharp corners=downhill,
    boxrule=0.5pt,
    left=6pt, right=6pt, top=6pt, bottom=6pt,
]
\ttfamily\small
You are a strict grader for one rubric criterion in a multimodal model evaluation.

\medskip
You will receive the original prompt, expected output modalities, input-file metadata, and exactly one model-produced modality payload for the rubric being graded. Grade only the supplied rubric against only the supplied model-produced modality payload.

\medskip
Do not infer credit from missing modalities, descriptions of absent media, or any model response content that was not supplied to you. If the supplied payload is insufficient to satisfy the rubric, score it 0.0.

\medskip
Return only valid JSON with this schema:
\begin{center}
\texttt{\{"score": 0.0, "explanation": "brief reason"\}}
\end{center}

Scoring:
\begin{itemize}[leftmargin=*,nosep]
    \item 1.0 means the criterion is fully satisfied.
    \item 0.0 means the criterion is not satisfied.
    \item Use partial credit between 0.0 and 1.0 when the criterion is partly satisfied.
\end{itemize}
\end{tcolorbox}

\begin{tcolorbox}[
    enhanced,
    colback=gray!5,
    colframe=black!70,
    colbacktitle=black!70,
    coltitle=white,
    fonttitle=\bfseries,
    title=Rubric Judge User Prompt Template,
    breakable,
    sharp corners=downhill,
    boxrule=0.5pt,
    left=6pt, right=6pt, top=6pt, bottom=6pt,
]
\ttfamily\small\raggedright
\# Original prompt\\
\{prompt\_text\}

\medskip
\# Expected output modalities\\
\{expected\_modalities\}

\medskip
\# Input files\\
\{input\_files\}

\medskip
\# Rubric modality being graded\\
\{modality\}

\medskip
\# Rubric criterion\\
\{rubric\}

\medskip
\# Supplied model-produced payload\\
\{payload\_summary\}

\medskip
Evaluate only this rubric criterion using only the supplied \{modality\} payload. Return only JSON.
\end{tcolorbox}

\noindent Non-text artifacts are appended to the same call as attached media, each group introduced by a label so that the judge cannot confuse the artifact it is grading with the artifact the model was given.

\begin{tcolorbox}[
    enhanced,
    colback=gray!5,
    colframe=black!70,
    colbacktitle=black!70,
    coltitle=white,
    fonttitle=\bfseries,
    title=Attached Media Labels,
    breakable,
    sharp corners=downhill,
    boxrule=0.5pt,
    left=6pt, right=6pt, top=6pt, bottom=6pt,
]
\ttfamily\small\raggedright
\# Attached below: the model-produced \{modality\} artifact. This is the artifact to grade.

\medskip
\# Attached below: the artifact(s) the model was given as prompt input. Context only. Never grade these as model output.
\end{tcolorbox}

\section{Full modality heatmap}

\section{Rubric writing guidelines}
\label{app:rubric_guidelines}

Below we summarize the guidelines followed when authoring the \bnshort rubrics.
Annotators worked from the finalized prompt set: for each prompt they received the prompt text, its input assets, and the gold output modalities, and returned one \emph{rubric set}, a numbered list of criteria, each tagged with exactly one output modality and each stating one requirement a correct answer must satisfy.

\subsection*{Guiding principle}

A rubric is a grading instrument, not a specification of the answer.
It captures only those properties whose presence or absence settles correctness, one per criterion; it does not attempt to describe in full what the asset should look or sound like.
Two rules follow.

\begin{itemize}
    \item \textbf{Write for a blind, single-modality grader.}
    Each criterion is graded against the model's output for the one modality it is tagged with, and must be decidable from that output alone: without the model's output for the other modalities, and without the prompt or its input assets. A criterion that can only be settled by comparing two output modalities, or by inspecting an input, must not be written. (The grader is supplied the prompt and inputs as background context, but no criterion may depend on them.)
    \item \textbf{Settle the answer inside the criterion.}
    Never merely name a topic the response should address; state the fact, value, or property it must contain, so the grader never has to derive the correct answer or fall back on its own knowledge. \textit{Bad:} Identifies the breed of the dog. \textit{Good:} Identifies the dog as an Australian Shepherd or Miniature American Shepherd.
\end{itemize}

\noindent This also covers prompts that ask the model to edit or transform a supplied asset: the criterion names the property the output must have, stated concretely enough to check without the input (``The map in the output image is viewed from directly above'', not ``Preserves the perspective of the original'').

\subsection*{Procedure}

\begin{enumerate}[nosep]
    \item \textbf{Confirm the output modalities.} Decide which modalities a genuinely useful answer would need and compare against the gold list, reading for the user's underlying goal rather than for keywords; flag the gold list rather than writing against one you disagree with. Two conventions apply: modalities nested inside others are not labeled separately (a PDF containing diagrams is \emph{document} only, a video with a soundtrack is \emph{video} only), and \emph{text} is an output modality only when the prompt calls for textual content in its own right, not for commentary wrapped around a generated asset.
    \item \textbf{Partition the requirements.} Assign each requirement to the single modality that carries it. Every gold modality must receive criteria, each criterion is tagged with exactly one modality, and a modality's criteria should cover everything the prompt asks of that modality and nothing else. \textbf{Never grade one modality's criterion against another's content}: if a fact is required in both a diagram and a written explanation, assert it in both the image and the text criteria.
    \item \textbf{Write the criteria.} Order each modality's criteria as: one \emph{delivery} criterion asserting that the asset exists and is of the right kind, including sub-kind where specified (``Delivers a CSV document file.''), which text normally omits; for non-text assets an \emph{integrity} criterion where it is a plausible failure mode (the asset opens or plays without corruption, truncation, or rendering errors); then the \emph{content} criteria, which are the bulk of the rubric; then at most one or two \emph{craft} criteria on legibility, structure, or pacing, where usefulness genuinely depends on execution.
\end{enumerate}

\noindent Every criterion must be:

\begin{itemize}[nosep]
    \item \textbf{Self-contained and atomic}: it settles the expected answer itself and asserts one fact. ``Labels sunlight and water and carbon dioxide as inputs'' should be three criteria.
    \item \textbf{Mutually exclusive}: no two criteria award credit for the same content, which would silently double-weight it within its modality.
    \item \textbf{Declarative}: a third-person statement about what the output does (``Delivers\ldots'', ``Shows\ldots'', ``The diagram depicts\ldots''), never a question or an instruction to the model.
    \item \textbf{Binary-decidable and objectively verifiable}: answerable yes or no from the asset itself plus, where relevant, publicly checkable fact.
    \item \textbf{Timeless}: anchor anything time-dependent. \textit{Bad:} States the current marathon world record. \textit{Good:} States that the men's marathon world record as of January 2025 is 2:00:35.
    \item \textbf{Achievable in the modality}: do not ask an image to carry a causal argument only prose can express, or audio to be ``well formatted''.
    \item \textbf{Generation-agnostic}: specify properties of the output, never the technique or tool used to produce it.
    \item \textbf{Tolerant where the answer legitimately varies}: enumerate the acceptable set with ``such as'' or ``or'' rather than silently picking one.
\end{itemize}

\noindent\textbf{Subjectivity.}
Open-ended and creative prompts still need objective criteria: grade the observable properties that make the output fit for purpose, not the impression it creates.
Replace ``The image is visually appealing'' with ``Uses readable labels and an uncluttered visual layout that makes the inputs, process, and outputs easy to distinguish''.

\noindent\textbf{Common mistakes.}
Referring to the prompt or user from inside a criterion (``as requested''); requiring consistency across output modalities; grading exact wording of speech rather than its content; padding with criteria the prompt does not support; restating delivery further down the list; phrasing a requirement negatively where a positive statement would do.

\noindent\textbf{How many.}
Target \textbf{4--12 criteria per output modality}.
Generating substantial content from scratch warrants more, a small edit to a supplied asset fewer.
Scoring averages within a modality before averaging across modalities, so writing more criteria for a modality sharpens how finely it is measured without giving it more weight in the prompt's score, and a tight set of well-separated criteria scores far more reliably than a long list of overlapping ones.

\subsection*{Output format}

Return one JSON object per prompt, with criteria numbered from 1 across all modalities and \texttt{modality} lower-case and drawn from the prompt's gold output modalities.
All criteria within a modality carry equal weight, so there is no weight field: the score for a modality is the fraction of its criteria satisfied, and the prompt's score is the mean of those per-modality scores.

\begin{tcolorbox}[
    enhanced,
    colback=gray!5,
    colframe=black!70,
    colbacktitle=black!70,
    coltitle=white,
    fonttitle=\bfseries,
    title=Rubric Schema,
    breakable,
    sharp corners=downhill,
    boxrule=0.5pt,
    left=6pt, right=6pt, top=6pt, bottom=6pt,
]
\ttfamily\small\raggedright
\{"criteria": [\\
\hspace*{2em}\{"id": "1", "criterion": "<one requirement>", "modality": "<text|image|audio|video|document>"\},\\
\hspace*{2em}\ldots\\
]\}
\end{tcolorbox}

\subsection*{Worked example}

\noindent\textit{``Can you create a list of pros and cons between hybrid and fully electric cars and send it to me in a C S V file?''} (spoken audio input, \emph{document} output).
Criterion 1 pins the file type the prompt named, criterion 2 checks structure, and the content criteria enumerate acceptable answers rather than fixing one.

\begin{enumerate}[nosep,leftmargin=2em]
    \item Delivers a CSV document file.
    \item Formats the file as a valid tabular CSV with a header row and comma-separated rows.
    \item Includes at least two pros for hybrid cars, such as better fuel efficiency than gas-only cars, reduced range anxiety, or less reliance on charging stations.
    \item Includes at least two cons for hybrid cars, such as continued gasoline use, tailpipe emissions, more complex maintenance, or limited electric-only driving.
    \item Includes at least two pros for fully electric cars, such as zero tailpipe emissions, lower routine maintenance, home charging, or lower operating costs.
    \item Includes at least two cons for fully electric cars, such as charging time, charger availability, driving range limits, battery degradation, or higher upfront cost.
\end{enumerate}

\noindent The crochet rubric set in \cref{subsec:rubrics} is the multi-modality counterpart: each modality answers a different clause of the prompt, and the audio criterion carries the expected pronunciation itself rather than deferring to the video.

\section{Human annotation guidelines}
\label{app:annotation_guidelines}

The instructions below were given to the human annotators whose judgments we compare against the rubric-based LLM judge in \cref{subsec:rubric_evaluation}.
They are reproduced with light edits for formatting and to match the annotation task as it was finally delivered.

\subsection*{Instructions as given to annotators}

You will get a set of questions asked to a model and the model answers to those questions.
Those questions typically involve multiple modalities in both the input and the output.
Your task is to provide a per-modality assessment of whether the model's answer to the question is correct.
The modalities present in the inputs and outputs of the benchmark are: text, audio, image, video and document, where document can be any type of document (txt, pdf, docx, ppt, etc.).

\subsection*{Input / expected output}

\noindent\textbf{Input:}
\begin{enumerate}
    \item A question / prompt involving one or more modalities.
    \item The list of output modalities that are required to answer that question satisfactorily (the prompt's desired output modalities).
    \item A model-generated answer to this question.
\end{enumerate}

\noindent\textbf{Expected output:}
\begin{enumerate}
    \item For each of the prompt's desired output modalities, an assessment of the correctness of the model's response for that output modality: `correct' or `incorrect'.
    \item For each such assessment, a short free-text comment justifying the verdict.
\end{enumerate}

Judge every desired output modality, also if the model did not return anything for that modality.
Modalities that are not in the list are not judged.

\subsection*{Additional instructions}
\begin{itemize}
    \item If the model did not generate a specific output modality that was needed to answer the question, mark that output modality as `incorrect'.
    \item Do not provide a fine-grained judgment of correctness: the verdict is binary, `correct' or `incorrect'. Always add a short comment stating why, and use it as well to flag doubt or anything else worth recording.
    \item A link to an asset counts as returning that modality. If a model provides a link in its answer but the link is broken, you may mark its
    response for that modality incorrect.
    \item If a question requires aligning two modalities (e.g.\ audio and video) and the model correctly produces both but does not align them, you may choose one of the modalities to mark incorrect.
    \item Modalities \emph{within} other modalities are not marked separately.
    For example:
    \begin{itemize}
        \item If the user asks to return a pdf with instructions, the desired output modality is \textbf{document} only, not also text or image if the document contains texts or images.
        \item If the user asks to return a video and that video contains audio, audio is not a separate modality.
    \end{itemize}
    \item Text that merely accompanies another modality is never judged in itself; text is assessed only when it is one of the desired output modalities, in which case it is judged like any other modality (see the example below). Likewise, modalities the model produced but that were not asked for are not judged separately, though you may mention them in the comment.
\end{itemize}

\subsection*{Knowledge acquisition using AI / LLMs}
Some questions may require expert knowledge to be answered; you may use an AI model to help acquire that knowledge and indicate you did so for that question. However, you may not:
\begin{itemize}
    \item Pass the model answer to an AI or LLM and ask if it is correct.
    \item Pass the original question directly to an AI or LLM and use its response to judge the correctness of the provided model answer.
\end{itemize}

\subsection*{Example}

\noindent\textit{\textbf{Question}}\\
Can you show me how to make soft chocolate chip cookies? A list of ingredients and a tutorial video would be nice. Also give me an image of chocolate chip cookies that I can send to my friends to invite them to come eat the cookies.

\medskip
\noindent\textit{\textbf{Desired output modalities:}} text, video, image.

\medskip
\noindent\textit{\textbf{Model answer:}}\\
Sure, here is a video with a tutorial for how to make chocolate chip cookies: \url{https://youtu.be/PFJAuAWxuvI?si=_dqH10g3xEFOeEJz}.

\smallskip
\noindent List of ingredients:
\begin{itemize}
    \item Brown Sugar \& White Sugar
    \item Unsalted Butter \& Salt
    \item Large Egg
    \item Vanilla Extract
    \item All-Purpose Flour
\end{itemize}

\medskip
\noindent\textit{\textbf{Judgment:}}
\begin{itemize}
    \item \textit{Video}: correct. The model returned a video tutorial to make chocolate chip cookies as required.
    \item \textit{Text}: incorrect. The model did return a list of ingredients, but baking soda and chocolate chips were missing from the list.
    \item \textit{Image}: incorrect. The model did not return an image.
\end{itemize}

\section{\mmiscores for the rubric-validation runs}
\label{app:mmi_scores}

The experiment of \cref{subsec:rubric_eval_setup} was run to exercise the rubrics, not to rank models: these values document a different, later set of models run with Gemini generation backends, and are neither a ranking nor comparable to \cref{tab:overall_metrics}.
We record them here so that the run is fully documented and the numbers behind \cref{subsec:rubric_evaluation} can be checked.
\cref{tab:mmi_scores} gives, for each of the three tool-scaffolded models, the \mmiscore over all \nsamples prompts, the same score with document criteria removed, and the \mmiscore restricted to each modality in turn.
The scaffolding raises modality presence enough for the \mmiscore to reflect gradeable content, which is not the case for the main-evaluation models (\cref{subsec:mmi_score}).
No document-generation tool was provided, so the document column is zero by construction.

\begin{table}[h]
    \centering
    \begin{tabular}{l cc ccccc}
\toprule
& \multicolumn{2}{c}{\mmiscore} & \multicolumn{5}{c}{\mmiscore\ by modality} \\
\cmidrule(lr){2-3} \cmidrule(lr){4-8}
Model & All & Excl.\ doc. & Text & Image & Audio & Video & Doc. \\
\midrule
Gemini 3.5 Flash & 0.51 & 0.63 & 0.69 & 0.78 & 0.53 & 0.47 & 0.00 \\
Claude Opus 4.8 & 0.36 & 0.44 & 0.51 & 0.46 & 0.42 & 0.32 & 0.00 \\
GPT-5.4 & 0.29 & 0.35 & 0.59 & 0.39 & 0.20 & 0.22 & 0.00 \\
\bottomrule
\end{tabular}

    \caption{\textbf{\mmiscores for the three tool-scaffolded models of \cref{subsec:rubric_eval_setup}.}
    \emph{All} is the \mmiscore as defined in \cref{subsec:mmi_score}: the mean over the \nsamples prompts of each prompt's mean across its output modalities of the mean criterion score within that modality.
    \emph{Excl.\ doc.} is the same quantity with document criteria removed, over the prompts that retain at least one non-document output modality.
    The five modality columns apply that same within-modality mean, averaged over the prompts for which the modality in question is expected; because different combinations of modalities are expected across prompts, the columns do not average to the \mmiscore.
    Document is zero throughout because no document-generation tool was provided.}
    \label{tab:mmi_scores}
\end{table}

\end{document}